\documentclass[letterpaper, 10 pt, conference]{ieeeconf}  

\IEEEoverridecommandlockouts                              

\usepackage{graphics}
\usepackage{epsfig}
\usepackage{mathptmx}
\usepackage{times}
\usepackage{amsmath}
\usepackage{amssymb}
\usepackage{multirow}
\usepackage{cite}
\usepackage{url}
\makeatletter
\let\NAT@parse\undefined
\makeatother
\usepackage[
colorlinks=true,
linkcolor=blue,
urlcolor=blue,
citecolor=blue
]{hyperref}

\title{\LARGE \bf SurGPT: A Multi-Task Foundation Model for Robotic Surgical Skills Assessment and Gesture Recognition}

\author{Parham~M.~Kebria,
~Saeid~Nahavandi,
~Mehrdad~Saif,
~and~Fernando~Bello
\thanks{P. M. Kebria is with the Electrical and Computer Engineering Department, North Carolina A\&T State University, Greensboro, NC, USA, and also with Swinburne University of Technology, Hawthorn, VIC, Australia. (This work was done with Swinburne University of Technology.)
    }
\thanks{S. Nahavandi is with Swinburne University of Technology, Hawthorn, VIC, Australia, and also with Harvard Paulson School of Engineering and Applied Sciences, Harvard University, Allston, MA, USA.
    }
\thanks{M. Saif is with the Department of Electrical and Computer Engineering, University of Windsor, Windsor, Ont. Canada. 
    }
\thanks{F. Bello is with Imperial College London, London, United Kingdom, and also with Duke-NUS Medical School, Singapore.
    }
\thanks{Corresponding author: {\small pkebria@swin.edu.au}}
}

\begin{document}

\maketitle

\thispagestyle{empty}
\pagestyle{empty}

\begin{abstract}

The assessment of surgical skills in robotic surgery training has traditionally relied on subjective evaluations by human experts, creating bottlenecks in surgical education and limiting scalability. We present a novel foundation model that automatically analyzes robotic surgical videos to provide comprehensive skill assessment and gesture understanding. Combining a visual encoder with a transformer architecture, the model performs three key tasks simultaneously: 1) surgical task classification (knot tying, needle passing, suturing), 2) skill level assessment on a standardized scale, and 3) natural language generation of surgical gesture descriptions. The model processes video sequences from the da Vinci surgical robot and generates human-readable text descriptions of surgical actions, similar to how language models generate text from prompts, using a multi-task learning framework with adaptive uncertainty weighting to optimize performance across all objectives. Our transformer-based architecture incorporates positional encoding for temporal understanding and attention mechanisms for spatial-temporal feature integration. Evaluated on the da Vinci Skills Simulator dataset, the proposed framework demonstrates state-of-the-art performance in surgical task recognition and provides interpretable skill assessments that correlate with expert evaluations, with the model's ability to generate natural language descriptions of surgical gestures opening new possibilities for automated surgical education, real-time feedback systems, and objective skill assessment in surgical training and certification programs.

\end{abstract}


\section{Introduction}

The emergence of robotic systems in clinical and healthcare practices makes them a critical element in a variety of medical procedures \cite{freschi2013technical, kebria2023haptically, kebria2025hercules}. Therefore, surgeons and clinicians need to learn, adapt, and practice the new technology. Employing all their skills, knowledge, and experience in this new schema of practice becomes a challenge, considering their limited time for training. This phenomenon requires a methodology to properly evaluate surgical skills in robotic practices \cite{carstens2025artificial}. Surgical skill assessment in robotic surgery training has traditionally relied on subjective evaluations by human experts, creating bottlenecks in education and restricting scalability \cite{walshaw2025essential}. However, such expert-driven evaluations are time-intensive, costly, and inherently inconsistent, as they are susceptible to variability and human bias \cite{kebria2019deep,mattioli2025objective}.

Recent advances in deep learning have significantly expanded the scope of surgical scene understanding, shifting from handcrafted features and task-specific models toward end-to-end learning frameworks. In particular, transformer-based architectures have emerged as a powerful paradigm for modeling long-range temporal dependencies and complex spatiotemporal interactions in surgical videos \cite{kiyasseh2023vision}. By leveraging self-attention mechanisms, transformers can capture both local motion patterns and global procedural context, enabling a more holistic understanding of surgical scenes compared to traditional sequential models.

The foundation model paradigm enables transfer learning across different surgical procedures and institutions, potentially revolutionizing surgical education through scalable, objective, and consistent skill evaluation \cite{min2025innovating}. Toward this paradigm, a wide range of techniques have been investigated to utilize expert human demonstrations to provide data-driven models with the vast amount of knowledge and context of the target procedures \cite{kebria2021stable,zare2023leveraging,kebria2025learning}. Surgical scene understanding introduces advanced machine learning and deep learning techniques used in image processing in the context of surgical procedures. This includes instrument detection, instrument-tissue segmentation, instrument tracking, instrument pose estimation, and tissue classification, to name a few \cite{khan2025comprehensive}. Foundation models have shown promising performance in precision and effectiveness in the analysis of medical images and anatomical information \cite{zhang2024segment}.

Despite these advances, there are still drawbacks and challenges in the deployment of foundation models in clinical applications that have not yet been investigated. Being large in size with the corresponding large number of parameters, foundation models are extremely data-thirsty. However, the privacy of healthcare data is a major concern, which limits the accessibility of large enough medical datasets for training such models \cite{liu2025preserving}. On the other hand, the lack of interpretability and explainability in these deep models poses significant challenges to their adoption in clinically critical applications \cite{rudin2019stop}. Still, foundation models provide novel benefits in the field, such as integrating multi-modal data to obtain more comprehensive diagnosis decisions \cite{liu2025visual}. Specifically, extracting diagnostic information from clinical videos is one of the areas where foundation models demonstrate unmatched benefits \cite{schmidgall2024general}. 

Moreover, surgical videos are characterized by high variability due to differences in anatomy, surgeon style, viewpoint, and occlusions, while annotated datasets remain limited and costly to obtain. Specifically, for clinical adoption, surgical scene understanding systems must provide interpretable and reliable outputs that align with expert reasoning. Transformer-based foundation models, particularly when coupled with interpretable outputs such as natural language descriptions, represent a promising direction for addressing these challenges and advancing automated, scalable, and clinically meaningful surgical intelligence systems. Existing surgical AI methods are often designed for single tasks, which limits transfer across related clinical objectives. They also provide limited interpretability and rarely use uncertainty as an adaptive learning signal, reducing trust and robustness. 

 Motivated by prior limitations, we propose a unified multi-task foundation model for robotic surgical video analysis that jointly performs gesture recognition and real-time skill assessment, combining a visual encoder and transformer for shared spatiotemporal reasoning, improved interpretability, and adaptive uncertainty weighting for better generalization and reliability. The main contributions of this work are summarized as follows:
\begin{itemize}
\item Automated surgical task recognition: We propose a unified framework that accurately identifies core robotic surgical tasks, including knot tying, needle passing, and suturing, directly from video sequences.
\item Objective surgical skill assessment: The model provides a standardized and scalable evaluation of surgical skill levels, producing objective proficiency scores that correlate strongly with expert assessments.
\item Interpretable gesture understanding via natural language generation: We introduce a novel capability for generating human-readable textual descriptions of fine-grained surgical gestures from video data, improving interpretability and supporting automated feedback and educational applications.
\end{itemize}

 The model processes video sequences captured from the da Vinci surgical robot and produces human-readable textual explanations of surgical actions, analogous to prompt-driven language generation in large language models. A multi-task learning strategy with adaptive uncertainty weighting is employed to balance and optimize performance across all objectives. The transformer architecture incorporates positional encoding to capture temporal dynamics and attention mechanisms for effective spatiotemporal feature integration. Extensive evaluation on the da Vinci Skills dataset \cite{gao2014jhu,ahmidi2017jigsaws} demonstrates that the model achieves state-of-the-art performance in surgical task recognition, while providing interpretable skill assessments that closely correlate with expert evaluations. Moreover, its ability to generate natural language descriptions of surgical gestures introduces new opportunities for automated surgical education, real-time feedback, and objective skill assessment in robotic surgery training and certification programs.

\section{Surgical Scene Understanding}
Surgical scene understanding aims to computationally interpret the visual and temporal context of surgical procedures by recognizing tools, actions, phases, and skill-related patterns from intraoperative data. In robot-assisted and minimally invasive surgery, this problem is primarily addressed through the analysis of surgical videos and kinematic signals to enable context-aware assistance, objective skill assessment, and intelligent feedback systems \cite{carstens2025artificial}.


Surgical motion can be viewed as being similar to human language because it is made up of small, meaningful actions that occur in a structured sequence. These actions follow implicit rules and constraints, just as words and sentences do in language. As a result, methods originally developed for modeling human language and speech can also be applied to the analysis of surgical motion.


At Johns Hopkins University, the Language of Surgery project studies surgical activity as a form of complex and dexterous human motion. The project aims to build curated surgical motion datasets, establish protocols for secure data storage and sharing, and develop computational models to analyze surgical activity. These models are then used to support surgical training, objective skill assessment, and research on human–machine collaboration in surgery.

Surgical activity data include multiple sources of information, such as surgical tool kinematics, endoscopic video, system event logs, and additional signals such as surgeon posture, speech, and manual annotations. Together, these data provide a comprehensive representation of surgical behavior that supports systematic analysis and learning.

A central challenge in surgical scene understanding is the lack of standardized, fine-grained annotations that capture both the temporal structure and skill-related aspects of surgical activity. The JIGSAWS (JHU–ISI Gesture and Skill Assessment Working Set \cite{ahmidi2017jigsaws}) dataset was introduced to address this gap and has since become a foundational benchmark for modeling and evaluating surgical scene understanding in robot-assisted surgery.

\begin{figure}[b]
  \centering
  \includegraphics[width=\columnwidth]{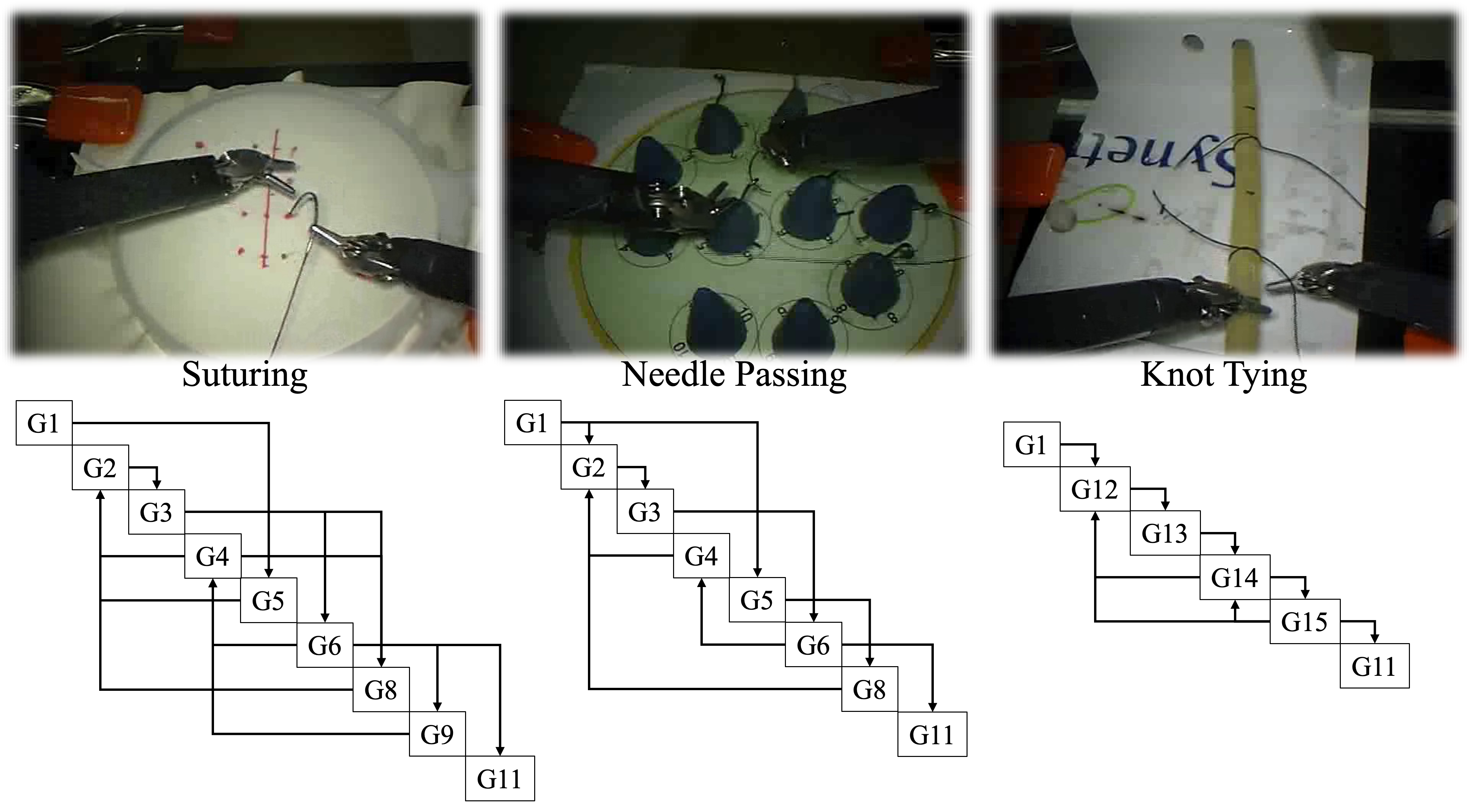}
  \caption{Example scene frames of the three surgical tasks of JIGSAWS and their corresponding task orders and flow of gestures, from Suturing (left) to Knot Tying (right).}
  \label{fig:jigsaws}
\end{figure}

\subsection{JIGSAWS Dataset}
JIGSAWS consists of synchronized video and kinematic recordings collected from the da Vinci surgical system during the execution of three fundamental robotic surgery tasks: knot tying, needle passing, and suturing (Fig. \ref{fig:jigsaws}). These tasks represent core psychomotor skills essential in robotic surgery and are commonly incorporated into surgical training curricula. The dataset comprises multiple trials conducted by surgeons with varying levels of expertise, categorized as novice, intermediate, and expert, to facilitate a systematic analysis of both task execution and surgical skill proficiency.

A defining characteristic of JIGSAWS is its fine-grained temporal annotation of surgical gestures, referred to as surgemes. Each trial is segmented into a sequence of elementary gestures that capture meaningful units of surgical motion and interaction. The gesture taxonomy includes eleven standardized classes, denoted as G1–G11, which describe recurrent actions across tasks:

G1: Reach for needle with right hand

G2: Position needle

G3: Push needle through tissue

G4: Transfer needle from right to left hand

G5: Move to center with needle in left hand

G6: Pull suture with left hand

G7: Pull suture with right hand

G8: Orient needle

G9: Use right hand to help tighten suture

G10: Use left hand to help tighten suture

G11: Drop suture at end of task

This gesture-level decomposition provides a structured representation of surgical activity, enabling models to learn the temporal organization and hierarchical composition of surgical motion. In addition to gesture annotations, JIGSAWS provides standardized skill scores based on expert evaluations, supporting objective assessment of surgical proficiency and correlation analysis between automated predictions and human judgment.


Despite its impact, JIGSAWS exposes several open challenges in surgical scene understanding. The dataset is relatively small by modern deep learning standards, and its annotations focus primarily on discrete gesture labels and numeric skill scores, without explicit semantic descriptions of surgical intent or context. Moreover, many existing methods treat gesture recognition, task classification, and skill assessment as separate problems, resulting in fragmented pipelines that fail to capture global procedural structure and limit interpretability. 

\subsection{Skill Scores in the JIGSAWS Dataset}
In JIGSAWS, surgical skill is quantified using expert-derived global rating scores, not gesture-level labels. These scores are provided per trial and are commonly used as ground truth for skill assessment and regression/classification tasks. Each surgical trial is evaluated by expert surgeons using a standardized Global Rating Scale (GRS) adapted for robotic surgery \cite{gao2014jhu}. The GRS consists of the following six skill metrics:
\begin{itemize}
    \item\textbf{Depth Perception}: Ability to judge spatial relationships and accurately manipulate tools in 3D space.
    \item\textbf{Bimanual Dexterity}: Effective and coordinated use of both hands during task execution.
    \item\textbf{Efficiency}: Economy of motion and avoidance of unnecessary or redundant movements.
    \item\textbf{Force Sensitivity}: Appropriate application of force to tissue and instruments.
    \item\textbf{Robotic Control}: Smoothness and precision of tool manipulation using the robotic interface.
    \item\textbf{Overall Skill}: Holistic assessment of surgical proficiency.
\end{itemize}
Each metric is scored on a Likert scale from 1 to 5, where \textbf{1} indicates poor performance, and \textbf{5} indicates expert-level performance. These scores are typically averaged across multiple expert raters to reduce subjectivity and are provided as numerical labels that support skill regression (predicting continuous scores), skill classification (e.g., novice vs. expert), and correlation analysis with automated assessment models. These properties make JIGSAWS both a benchmark and a motivating testbed for unified modeling approaches. Its hierarchical temporal annotations, multimodal structure, and standardized skill labels provide an ideal foundation for developing models that move beyond isolated predictions toward holistic surgical scene understanding. In the following section, we build upon these characteristics to introduce a transformer-based foundation model that jointly models spatiotemporal dynamics, surgical task structure, and skill-related information within a single end-to-end framework.


\section{Methodology}
\label{sec:methodology}

This section presents the comprehensive transformer-based foundation model developed for surgical scene understanding, encompassing task classification, skill assessment, and scene description generation.

\subsection{Model Architecture Overview}

The developed framework implements a multi-modal transformer-based foundation model that combines video understanding with natural language processing for comprehensive surgical scene analysis. The architecture consists of three main components: (1) a spatial-temporal feature extraction pipeline, (2) a transformer-based sequence modeling backbone, and (3) multi-task prediction heads for simultaneous skill assessment and scene description.

\subsection{Video Encoder Architecture}

\subsubsection{Spatial Feature Extraction}
The framework utilizes pre-trained convolutional neural networks (VGG-16/19 or ResNet-18/50) as backbone encoders for extracting spatial features from surgical video frames. The backbone networks, pre-trained on ImageNet, are modified by removing the final classification layers and adding adaptive pooling to generate fixed-size $7 \times 7$ spatial feature maps. For input video $\mathbf{V} \in \mathbb{R}^{B \times T \times C \times H \times W}$ with batch size $B$, temporal sequence length $T$, channels $C$, height $H$, and width $W$:
\begin{equation}
\mathbf{F}_{spatial} = \text{AdaptivePool}(\text{CNN}_{\theta}(\mathbf{V}))
\end{equation}
where $\text{CNN}_{\theta}$ represents the backbone network and the output is reshaped to $(B \times T, 512 \times 7 \times 7)$ for VGG or $(B \times T, 2048 \times 7 \times 7)$ for ResNet.

\subsubsection{Feature Projection}
Spatial features are flattened and projected through a two-layer MLP with ReLU activation and dropout regularization:
\begin{equation}
\mathbf{F}_{proj} = \text{Dropout}(\text{ReLU}(\text{Linear}(\text{Flatten}(\mathbf{F}_{spatial}))))
\end{equation}
The projected features are then reshaped back to $(B, T, d_{feature})$ where $d_{feature} = 512$ is the feature dimension.

\subsection{Temporal Transformer Encoder}

\subsubsection{Positional Encoding}
To capture temporal relationships in surgical video sequences, we apply sinusoidal positional encodings:
\begin{align}
\text{PE}_{(pos,2i)} &= \sin\left(\frac{pos}{10000^{2i/d_{model}}}\right) \\
\text{PE}_{(pos,2i+1)} &= \cos\left(\frac{pos}{10000^{2i/d_{model}}}\right)
\end{align}
$pos$ is the temporal position and $i$ is the dimension index.

\subsubsection{Multi-Head Self-Attention}
The transformer encoder processes temporal sequences using multi-head self-attention with 8 attention heads:
\begin{equation}
\text{Attention}(\mathbf{Q}, \mathbf{K}, \mathbf{V}) = \text{softmax}\left(\frac{\mathbf{Q}\mathbf{K}^T}{\sqrt{d_k}}\right)\mathbf{V}
\end{equation}
\begin{equation}
\text{MultiHead}(\mathbf{Q}, \mathbf{K}, \mathbf{V}) = \text{Concat}(\text{head}_1, ..., \text{head}_h)\mathbf{W}^O
\end{equation}
where $\mathbf{W}^O \in \mathbb{R}^{h \cdot d_k \times d_{model}}$ is the learned output projection matrix that transforms the concatenated multi-head outputs back to the model dimension, with $h = 8$ attention heads and $d_k = d_{model}/h = 64$. The encoder consists of 6 transformer layers with $d_{model} = 512$, feed-forward dimension $d_{ff} = 2048$, and dropout rate $\delta = 0.1$.

\subsection{Multi-Task Learning Framework}
Here, we explain the three task-specific prediction heads for task classification, skill assessment, and gesture description, respectively.
\subsubsection{Task Classification Head}
A global pooling layer aggregates temporal features, followed by a two-layer MLP:
\begin{align}
&\mathbf{h}_{global} = \frac{1}{T}\sum_{t=1}^{T}\mathbf{h}_t\label{eq:task_head} \\
&\mathbf{y}_{task} = \text{Linear}(\text{ReLU}(\text{Linear}(\mathbf{h}_{global})))\label{eq:task_cls}
\end{align}

\subsubsection{Skill Assessment Head}
Implements both ordinal regression and categorical classification:
\begin{align}
&\mathbf{y}_{skill} = \text{Linear}(\text{ReLU}(\text{Linear}(\mathbf{h}_{global})))\label{eq:skill_head} \\
&\text{Expected Skill} = \sum_{i=1}^{5} i \cdot \text{softmax}(\mathbf{y}_{skill})_i\label{eq:skill_class}
\end{align}

\subsubsection{Gesture Text Generation Head}
Employs a transformer decoder with 6 layers for autoregressive text generation:
\begin{equation}\label{eq:text_head}
\mathbf{y}_{text} = \text{Decoder}(\mathbf{h}_{encoded}, \mathbf{x}_{text})
\end{equation}
where $\mathbf{x}_{text}$ represents input gesture tokens during training (teacher forcing) or just-generated tokens during inference.

\subsection{Adaptive Multi-Task Loss Function}

The framework implements uncertainty-based adaptive weighting for multi-task optimization, following the approach of Kendall et al.~\cite{kendall2018multi}. The total loss combines four task-specific components:
\begin{equation}\label{eq:total_loss}
\mathcal{L}_{total} = \sum_{i} w_i \mathcal{L}_i + \frac{1}{2}\log(\sigma_i^2)
\end{equation}
where $w_i = \frac{1}{\sigma_i^2}$ are learnable uncertainty weights and $\sigma_i^2$ represent uncertainties specific to the task.

\subsubsection{Individual Loss Components}

\paragraph{Task Classification Loss}
\begin{equation}\label{eq:task_loss}
\mathcal{L}_{task} = -\sum_{c=1}^{C} y_c \log(\hat{y}_c)
\end{equation}
in which $\hat{y}_c$ is the predicted task class from \eqref{eq:task_cls}, and the ground truth $y_c$ across all tasks $c\in\{1,\dots,C\}$.

\paragraph{Skill Assessment Loss (Combined)}
\begin{equation}\label{eq:skill_loss}
\mathcal{L}_{skill} = \alpha \mathcal{L}_{CE} + (1-\alpha) \mathcal{L}_{MSE}
\end{equation}
where $\mathcal{L}_{CE}$ is the cross-entropy for categorical prediction and $\mathcal{L}_{MSE}$ is the mean squared error for expected skill level regression with respect to \eqref{eq:skill_class} and \eqref{eq:skill_head}, respectively.

\begin{figure}[h]
    \centering
    \includegraphics[width=.99\columnwidth]{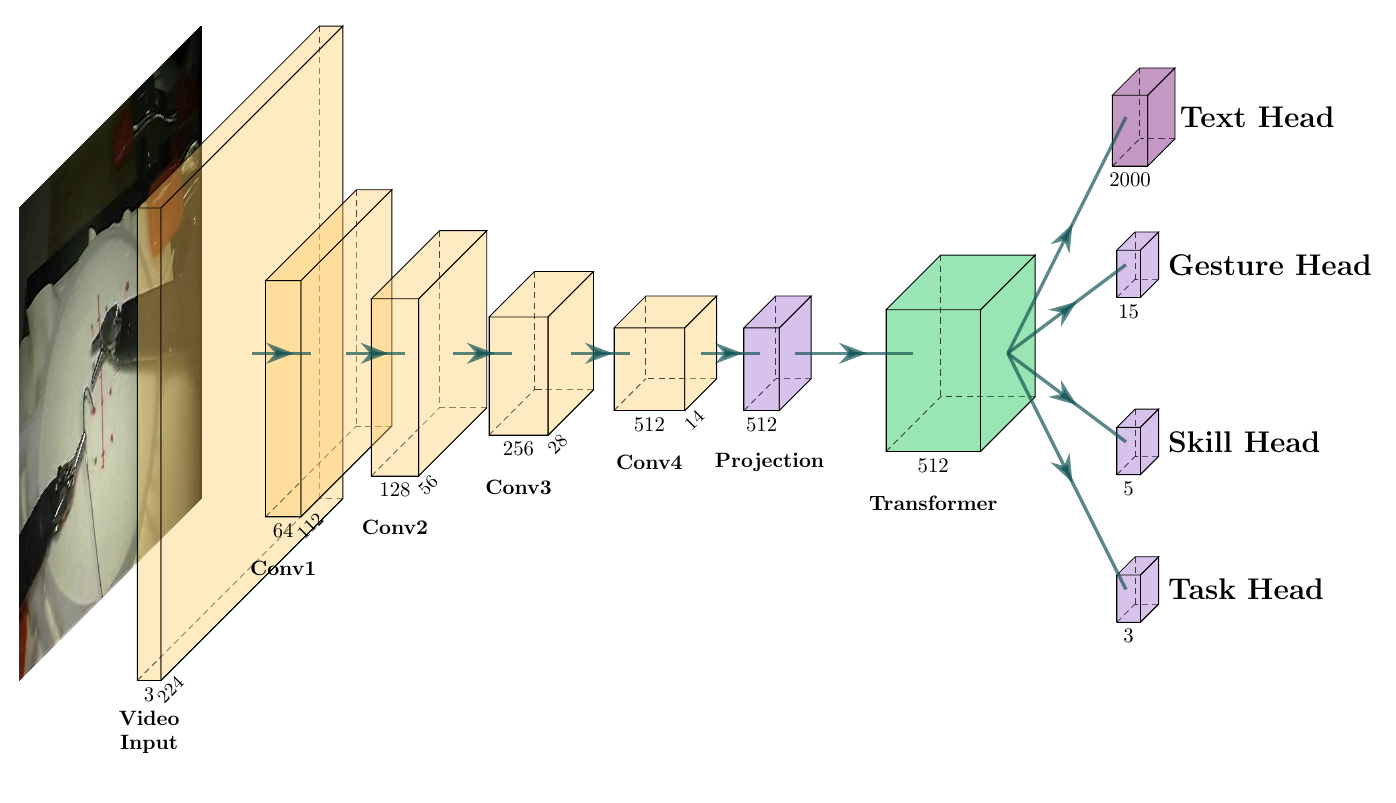}
    \caption{Multi-task transformer architecture for surgical skill assessment; a CNN backbone, temporal modeling, and specialized prediction heads.}
    \label{fig:arch}
\end{figure}

\paragraph{Text Generation Loss}
\begin{equation}\label{eq:text_loss}
\mathcal{L}_{text} = -\sum_{t=1}^{T_{text}} \log P(w_t | w_{1:t-1}, \mathbf{h}_{encoded})
\end{equation}
which is the log-likelihood $P$ of the next word $w_t$ given the previous sequence $w_{t-1}$ by $\mathbf{h}_{encoded}$ from \eqref{eq:text_head} for the entire text $T_{text}$. Figure \ref{fig:arch} depicts the schematic architecture of the proposed transformer-based multi-task model for surgical skill assessment. 

\subsection{Training Strategy}

The model training follows a two-phase transfer learning protocol. Initially, the pre-trained CNN backbone parameters are frozen for 10 epochs to stabilize the transformer components, followed by end-to-end fine-tuning of the entire network. Optimization is performed using AdamW with a learning rate of $5 \times 10^{-5}$, weight decay of $1 \times 10^{-5}$, and cosine learning rate scheduling with 20-epoch warmup. Gradient clipping at norm 0.5 prevents training instability, while mixed precision training with GradScaler enhances computational efficiency.

Data augmentation strategies include horizontal flipping (50\% probability), brightness and contrast adjustments within 0.8-1.2 range, and temporal downsampling (every 4th frame) to improve model robustness and generalization.

\subsection{Dataset and Evaluation}

\subsubsection{Data Structure}
The framework processes the da Vinci Skills Simulator dataset containing three surgical tasks: Knot Tying (fundamental suturing technique), Needle Passing (precision manipulation task), and Suturing (complete suturing procedure). Each video includes RGB video streams (30 FPS, downsampled to 10 FPS), kinematic data from robotic instruments, skill level annotations (1-5 scale), and gesture descriptions for text generation.

\subsubsection{Evaluation Metrics}

Performance evaluation employs task-specific metrics: accuracy, F1-macro, and F1-micro scores for task classification; Mean Absolute Error (MAE), Root Mean Square Error (RMSE), Pearson correlation coefficient (R²), and classification accuracy for skill assessment; and BLEU-4 score, ROUGE-L, and semantic similarity metrics for text generation quality.

\subsection{Model Complexity and Efficiency}

The foundation model contains approximately 25M parameters distributed across the video encoder (VGG-16:$\sim$14.7M), transformer backbone ($\sim$8.5M), and multi-task heads ($\sim$1.8M). The framework achieves real-time inference at $>$30 FPS on modern GPUs with input resolution of $224\times224$ pixels, maximum sequence length of 256 frames, vocabulary size of 2000 tokens, batch size of 16 sequences, and training for 100 epochs with early stopping (patience=15), enabling practical deployment in surgical training environments.

This comprehensive foundation model architecture enables unified surgical scene understanding through joint learning of task classification, skill assessment, and natural language generation, providing a robust framework for objective surgical training evaluation.

\section{Results and Discussions}
\label{sec:results}

This section presents the comprehensive evaluation results of the surgical skills foundation model across all three primary tasks: task classification, skill assessment, and surgical scene description generation.

Table~\ref{tab:overall_results} summarizes performance across all tasks, demonstrating the effectiveness of the multi-task learning framework. Presented in Table \ref{tab:task_classification}, the surgical task classification component achieved perfect performance in all evaluation metrics, correctly identifying all surgical procedures in the test set. Moreover, the skill assessment component demonstrated strong performance in predicting surgical skill levels, with a high correlation between the predicted and ground truth assessments as presented in Table \ref{tab:skill_assessment}. The surgical scene description generation component achieved high-quality natural language outputs, with strong performance across multiple evaluation metrics, as shown in Table \ref{tab:text_generation}.

\subsection{Performance Analysis}

\begin{figure}[htbp]
    \centering
    \includegraphics[width=0.99\columnwidth]{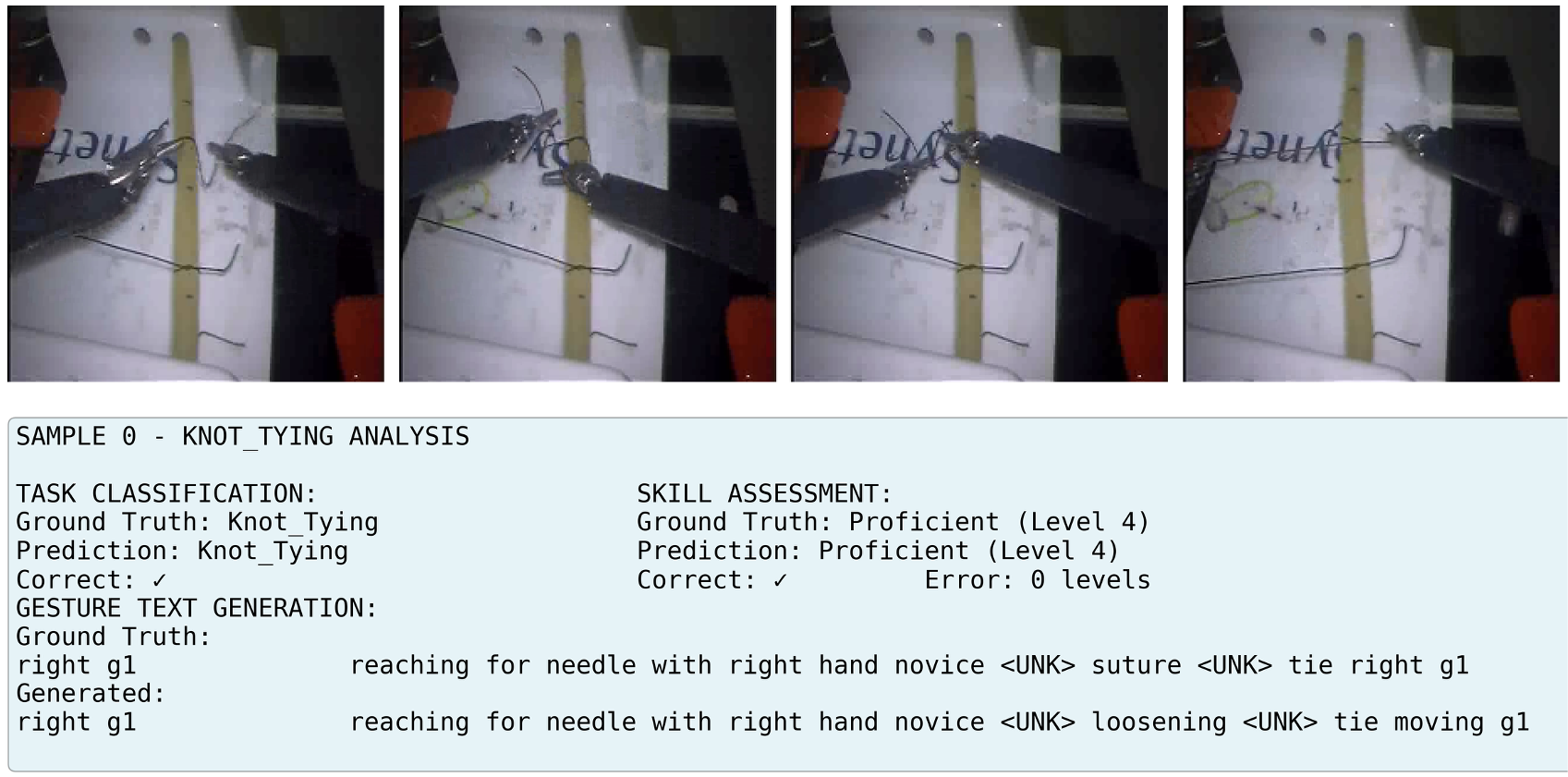}
    \caption{Sample multi-task performance of the model in a test scenario. The specific sample is presented to show the real evaluation of the model, where the generated text does not perfectly match the ground truth description.}
    \label{fig:sample}
\end{figure}

The experimental results demonstrate the effectiveness of the proposed multi-task learning framework. Perfect performance indicates excellent capability for surgical procedure identification, with the model successfully distinguishing between Knot Tying, Needle Passing, and Suturing tasks. Strong correlation metrics (Pearson: 0.833, Spearman: 0.904) demonstrate reliable skill level prediction, with the model achieving 91.7\% classification accuracy and low regression errors (MAE: 0.314, RMSE: 0.840). High BLEU-4 (0.847) and ROUGE-1 (0.962) scores indicate the generation of high-quality, relevant surgical scene descriptions with good semantic and syntactic alignment to ground truth annotations. The overall score of 92.1\% reflects the balanced performance across all tasks, validating the multi-task learning approach for comprehensive surgical scene understanding. Figure \ref{fig:sample} demonstrates a sample test scenario of the multi-task model.

\begin{table}[htbp]
\centering
\caption{Overall Performance Summary of the Surgical Skills Foundation Model}
\label{tab:overall_results}
\resizebox{.75\columnwidth}{!}{
{\footnotesize
\begin{tabular}{lcc}
\hline
\textbf{Task} & \textbf{Metric} & \textbf{Value} \\
\hline
\multirow{3}{*}{Task Classification} & Accuracy & 100.0\% \\
                                    & F1-macro & 100.0\% \\
                                    & F1-micro & 100.0\% \\
\hline
\multirow{4}{*}{Skill Assessment}   & MAE & 0.314 \\
                                   & RMSE & 0.840 \\
                                   & Pearson R² & 0.660 \\
                                   & Classification Accuracy & 91.7\% \\
\hline
\multirow{3}{*}{Text Generation}    & BLEU-4 & 0.847 \\
                                   & ROUGE-1 & 0.962 \\
                                   & ROUGE-2 & 0.895 \\
\hline
\multicolumn{2}{l}{\textbf{Overall Score}} & \textbf{92.1\%} \\
\hline
\end{tabular}}}
\end{table}


\begin{table}[htbp]
\centering
\caption{Detailed Task Classification Results}
\label{tab:task_classification}
{\footnotesize
\begin{tabular}{lc}
\hline
\textbf{Metric} & \textbf{Value} \\
\hline
Accuracy & 100.0\% \\
F1-macro & 100.0\% \\
F1-micro & 100.0\% \\
F1-weighted & 100.0\% \\
Precision (macro) & 100.0\% \\
Recall (macro) & 100.0\% \\
AUC (macro) & 100.0\% \\
AUC (weighted) & 100.0\% \\
\hline
\multicolumn{2}{l}{\textbf{Per-Task F1 Scores:}} \\
\hline
Knot Tying & 100.0\% \\
Needle Passing & 100.0\% \\
Suturing & 100.0\% \\
\hline
\end{tabular}}
\end{table}


\begin{table}[htbp]
\centering
\caption{Skill Assessment Evaluation Results}
\label{tab:skill_assessment}
\resizebox{.65\columnwidth}{!}{
{\footnotesize
\begin{tabular}{lc}
\hline
\textbf{Metric} & \textbf{Value} \\
\hline
Classification Accuracy & 91.7\% \\
F1-macro & 93.8\% \\
F1-weighted & 91.3\% \\
\hline
\multicolumn{2}{l}{\textbf{Regression Metrics:}} \\
Mean Absolute Error (MAE) & 0.314 \\
Root Mean Square Error (RMSE) & 0.840 \\
R-squared (R²) & 0.660 \\
Pearson Correlation & 0.833 \\
Pearson p-value & $4.45 \times 10^{-7}$ \\
Spearman Correlation & 0.904 \\
Spearman p-value & $1.33 \times 10^{-9}$ \\
\hline
\multicolumn{2}{l}{\textbf{Tolerance-based Accuracy:}} \\
Exact Match (±0) & 91.7\% \\
±1 Level Tolerance & 91.7\% \\
±2 Level Tolerance & 91.7\% \\
Adjacent Level Accuracy & 91.7\% \\
\hline
\end{tabular}}}
\end{table}


\begin{table}[htbp]
\centering
\caption{Text Generation Quality Assessment}
\label{tab:text_generation}
\resizebox{.6\columnwidth}{!}{
{\footnotesize
\begin{tabular}{lc}
\hline
\textbf{Metric} & \textbf{Value} \\
\hline
\multicolumn{2}{l}{\textbf{BLEU Scores:}} \\
BLEU-1 & 0.906 \\
BLEU-2 & 0.881 \\
BLEU-3 & 0.863 \\
BLEU-4 & 0.847 \\
\hline
\multicolumn{2}{l}{\textbf{ROUGE Scores:}} \\
ROUGE-1 (F-score) & 0.962 \\
ROUGE-1 (Precision) & 0.969 \\
ROUGE-1 (Recall) & 0.957 \\
ROUGE-2 (F-score) & 0.895 \\
ROUGE-L (F-score) & 0.937 \\
\hline
\multicolumn{2}{l}{\textbf{Additional Metrics:}} \\
Exact Match & 50.0\% \\
Token Accuracy & 0.13\% \\
Vocabulary Overlap & 91.9\% \\
Average Prediction Length & 55.0 tokens \\
Average Target Length & 51.3 tokens \\
Length Ratio & 1.074 \\
\hline
\end{tabular}}}
\end{table}

\section{Conclusions}
This work presented a comprehensive transformer-based foundation model for surgical scene understanding that successfully integrates task classification, skill assessment, and natural language description in a unified multi-task framework, achieving strong performance across all evaluation metrics with an overall score of 92.1\%, perfect task classification accuracy, reliable skill assessment with high correlation coefficients, and high-quality text generation. The framework's primary limitations include its scope, being limited to three basic surgical tasks on a single robotic platform, which limits generalizability to diverse surgical procedures and environments. Future work should focus on expanding the framework to more complex procedures across multiple surgical platforms while incorporating uncertainty-aware methods to improve trustworthiness and clinical applicability.

\bibliographystyle{IEEEtran}
\bibliography{surgptrefs}

@article{zare2023leveraging,
  title={Leveraging optimal transport for enhanced offline reinforcement learning in surgical robotic environments},
  author={Zare, Maryam and Kebria, Parham M and Khosravi, Abbas},
  journal={arXiv preprint arXiv:2310.08841},
  pages={1--9},
  year={2023}
}

@inproceedings{kebria2025learning,
  title={Learning Human-Robot Interactions in Perturbed Teleoperations},
  author={Kebria, Parham M and Nahavandi, Saeid and Howe, Robert D},
  booktitle={2025 IEEE 21st International Conference on Automation Science and Engineering (CASE)},
  pages={1987--1993},
  year={2025},
  organization={IEEE}
}

@article{kebria2025hercules,
  title={HERCULES: Haptically-Enabled Remotely Controlled Ultrasound Examination System},
  author={Kebria, Parham M and others},
  journal={IEEE Access},
  year={2025},
  volume={13},
  pages={75585--75598},
  publisher={IEEE}
}

@inproceedings{kebria2023haptically,
  title={Haptically-enabled robotic teleoperation for transcranial magnetic stimulation (TeleTMS)},
  author={Kebria, Parham M and others},
  booktitle={2023 IEEE International Conference on Systems, Man, and Cybernetics (SMC)},
  pages={2100--2105},
  year={2023},
  organization={IEEE}
}

@article{carstens2025artificial,
  title={Artificial intelligence for surgical scene understanding: a systematic review and reporting quality meta-analysis},
  author={Carstens, Matthias and others},
  journal={NPJ Digital Medicine},
  year={2025},
  volume={9},
  number={59},
  pages={1--11},
  publisher={Nature Publishing Group UK London}
}

@article{walshaw2025essential,
  title={Essential components and validation of multi-specialty robotic surgical training curricula: a systematic review},
  author={Walshaw, Josephine and others},
  journal={International Journal of Surgery},
  volume={111},
  number={4},
  pages={2791--2809},
  year={2025},
  publisher={LWW}
}

@article{mattioli2025objective,
  title={Objective Fluoroscopic Image-Based Assessment of Intraoperative Wire Navigation Skill Agrees with Subjective Expert Opinion},
  author={Mattioli, Dominik and others},
  journal={The Iowa orthopaedic journal},
  volume={45},
  number={1},
  pages={49},
  year={2025}
}

@article{min2025innovating,
  title={Innovating robot-assisted surgery through large vision models},
  author={Min, Zhe and Lai, Jiewen and Ren, Hongliang},
  journal={Nature Reviews Electrical Engineering},
  pages={350--363},
  year={2025},
  volume={2},
  publisher={Nature Publishing Group UK London}
}

@article{kebria2019deep,
  title={Deep imitation learning for autonomous vehicles based on convolutional neural networks},
  author={Kebria, Parham M and Khosravi, Abbas and Salaken, Syed Moshfeq and Nahavandi, Saeid},
  journal={IEEE/CAA Journal of Automatica Sinica},
  volume={7},
  number={1},
  pages={82--95},
  year={2019},
  publisher={IEEE}
}

@article{kebria2021stable,
  title={Stable neural adaptive filters for teleoperations with uncertain delays},
  author={Kebria, Parham M and Khosravi, Abbas and Nahavandi, Saeid},
  journal={IEEE Robotics and Automation Letters},
  volume={6},
  number={4},
  pages={8663--8670},
  year={2021},
  publisher={IEEE}
}

@article{freschi2013technical,
  title={Technical review of the da Vinci surgical telemanipulator},
  author={Freschi, Cinzia and others},
  journal={The International Journal of Medical Robotics and Computer Assisted Surgery},
  volume={9},
  number={4},
  pages={396--406},
  year={2013},
  publisher={Wiley Online Library}
}

@article{khan2025comprehensive,
  title={A comprehensive survey of foundation models in medicine},
  author={Khan, Wasif and others},
  journal={IEEE Reviews in Biomedical Engineering (Early Access)},
  year={2025},
  pages={1--22},
  publisher={IEEE}
}

@article{zhang2024segment,
  title={Segment anything model for medical image segmentation: Current applications and future directions},
  author={Zhang, Yichi and Shen, Zhenrong and Jiao, Rushi},
  journal={Computers in Biology and Medicine},
  volume={171},
  pages={108238},
  year={2024},
  publisher={Elsevier}
}

@article{liu2025preserving,
  title={Preserving privacy in healthcare: A systematic review of deep learning approaches for synthetic data generation},
  author={Liu, Yintong and Acharya, U Rajendra and Tan, Jen Hong},
  journal={Computer Methods and Programs in Biomedicine},
  volume={260},
  pages={108571},
  year={2025},
  publisher={Elsevier}
}

@article{rudin2019stop,
  title={Stop explaining black box machine learning models for high stakes decisions and use interpretable models instead},
  author={Rudin, Cynthia},
  journal={Nature machine intelligence},
  volume={1},
  number={5},
  pages={206--215},
  year={2019},
  publisher={Nature Publishing Group UK London}
}

@article{liu2025visual,
  title={Visual--language foundation models in medicine},
  author={Liu, Chunyu and others},
  journal={The Visual Computer},
  volume={41},
  number={4},
  pages={2953--2972},
  year={2025},
  publisher={Springer}
}

@article{schmidgall2024general,
  title={General surgery vision transformer: A video pre-trained foundation model for general surgery},
  author={Schmidgall, Samuel and Kim, Ji Woong and Jopling, Jeffrey and Krieger, Axel},
  journal={arXiv preprint arXiv:2403.05949},
  year={2024}
}

@inproceedings{gao2014jhu,
  title={JHU-ISI gesture and skill assessment working set (jigsaws): A surgical activity dataset for human motion modeling},
  author={Gao, Yixin and others},
  booktitle={MICCAI workshop: M2cai},
  volume={3},
  pages={3},
  year={2014}
}

@ARTICLE{ahmidi2017jigsaws,
  author={Ahmidi, Narges and others},
  journal={IEEE Transactions on Biomedical Engineering}, 
  title={A Dataset and Benchmarks for Segmentation and Recognition of Gestures in Robotic Surgery}, 
  year={2017},
  volume={64},
  number={9},
  pages={2025-2041},
  doi={10.1109/TBME.2016.2647680}
}

@article{kiyasseh2023vision,
  title={A vision transformer for decoding surgeon activity from surgical videos},
  author={Kiyasseh, Dani and others},
  journal={Nature biomedical engineering},
  volume={7},
  number={6},
  pages={780--796},
  year={2023},
  publisher={Nature Publishing Group UK London}
}

@inproceedings{kendall2018multi,
  title={Multi-task learning using uncertainty to weigh losses for scene geometry and semantics},
  author={Kendall, Alex and Gal, Yarin and Cipolla, Roberto},
  booktitle={Proceedings of the IEEE conference on computer vision and pattern recognition},
  pages={7482--7491},
  year={2018}
}







\end{document}